# Linguistically Grounded Models of Language Change

**Thierry Poibeau (thierry.poibeau@lipn.univ-paris13.fr)**
Laboratoire d'Informatique de Paris-Nord – CNRS and Université Paris 13 – 99 avenue Jean-Baptiste Clément
93430 Villetaneuse, France

## Introduction

Questions related to the evolution of language have recently known an impressive increase of interest (Briscoe, 2002). This short paper aims at questioning the scientific status of these models and their relations to attested data. We show that one cannot directly model non-linguistic factors (*exogenous factors*) even if they play a crucial role in language evolution. We then examine the relation between linguistic models and attested language data, as well as their contribution to cognitive linguistics.

## Endogenous and exogenous factors of language evolution

Languages are complex systems, whose evolution is due to a large number of exogenous and endogenous factors, similar to complex ecological entities (Haugen, 1972) (Mulhausler, 1996).

*Exogenous factors* – Exogenous factors include historical, political, social and geographical facts. For example, during invasions, dominant tribes may impose their language upon their neighbourhood; People willing to integrate another community may use the other community language; isolated languages like Icelandic tend to be more stable than contact languages; etc.

*Endogenous factors* – Part of a language may undergo profound changes due to analogy and linguistic instability. Following the influence of exogenous factors, linguistic structures may become instable, which can bring new dynamics in the evolution of a linguistic system, independently from any external influence. Phonetic evolution is full of such examples, where the evolution of a single sound makes the whole system unstable and quickly suffers major changes.

Among these factors, exogenous ones have a major impact on linguistic evolution. They can bring profound changes in a very short period of time. However, they are rarely taken into considerations by evolution models since these factors are hard to model, if even possible (the same phenomenon can be observed for economic models). Most language models are inspired by population models (Livingstone, 2003): they represent the context as a set of random variables, since context is for a large part unpredictable.

Thus, language evolution models are largely abstract models. Even if they provide an aid to understand the weight of different factors in language evolution, they are not prospective (they are not intended to predict the future of current languages) and can hardly be proven to be correct or wrong, except if they were applied to larger sets of data.

## Linguistically grounded models

Several studies have already stressed the need for more realistic models, based on attested facts. Previous studies in Historical linguistics may give a part of these data and provide a way to better validate existing models. Compared to existing models, grounding studies on attested facts is then crucial, since facts integrate external evolution causes. This domain is still at its very beginning (see Niyogi, 2002 for an application to the change of English word order).

Rastier (1999) presents a study in diachronic lexical semantics, using a morpho-dynamic model (the study concerns the evolution of the meaning of the noun *face* in French since the 15$^{th}$ century). The author demonstrates that "semantics, be it cognitive or not, can only have an anthropological foundation, articulated upon ethnology and history". For example, the precise study of the semantics of words through history allows determining how semantic prototypes are born, grow and disappear in a given language. The same kind of approach can be applied to other parts of linguistics, to phonology as well as to syntax. Frameworks like Optimality Theory or Game Theory (Jaeger, 2003) can easily model attested facts. Previous work in Historical Linguistics may give a part of these data and provide a way to better validate linguistic hypotheses.